\documentclass[runningheads]{llncs}

\usepackage{multirow}
\usepackage{colortbl}
\usepackage{eccv}

\usepackage{eccvabbrv}

\usepackage{graphicx}
\usepackage{booktabs}

\usepackage[accsupp]{axessibility} 

\usepackage{hyperref}

\usepackage{orcidlink}

\begin{document}

\title{MVP: Multimodal Emotion Recognition based on Video and Physiological Signals} 

\titlerunning{Multimodal for Video and Physio}

\author{\centering
Valeriya Strizhkova\inst{1} \and Hadi Kachmar\inst{1} \and Hava Chaptoukaev\inst{2}\and\\
Raphael Kalandadze\inst{3}\and Natia Kukhilava\inst{3}\and Tatia Tsmindashvili\inst{3}\and\\
Nibras Abo-Alzahab\inst{1}\and Maria A. Zuluaga\inst{2}\and Michal Balazia\inst{1}\and\\
Antitza Dantcheva\inst{1}\and Fran\c cois Br\'emond\inst{1}\and Laura M. Ferrari\inst{1,4}
}
    
\authorrunning{V.~Strizhkova et al.}

\institute{
Centre Inria d'Université Côte d'Azur, Sophia Antipolis, France 
\and
EURECOM, Sophia Antipolis, France
\and
Georgian Technical University, Tbilisi, Georgia
\and
Scuola Superiore Sant’Anna, Pontedera, Italy.
}

\maketitle

\begin{abstract}
Human emotions entail a complex set of behavioral, physiological and cognitive changes. Current state-of-the-art models fuse the behavioral and physiological components using classic machine learning, rather than recent deep learning techniques. We propose to fill this gap, designing the Multimodal for Video and Physio (MVP) architecture, streamlined to fuse video and physiological signals. Differently then others approaches, MVP exploits the benefits of attention to enable the use of long input sequences (1-2 minutes). We have studied video and physiological backbones for inputting long sequences and evaluated our method with respect to the state-of-the-art. Our results show that MVP outperforms former methods for emotion recognition based on facial videos, EDA, and ECG/PPG. The code is available on GitHub\footnote{\url{https://github.com/EmotionLab/EmotionVMAE}}.
\end{abstract}

\section{Introduction}
\label{sec:intro}

The recognition of human behavior from video data has been deeply investigated in the affective computing area~\cite{kahou2015recurrent}. 
What is limiting the performance of computer vision algorithms is the recognizability and the meaning of visual patterns. Facial expressions visible from cameras do not always correlate with the experienced emotions, which can lead to incorrect emotion recognition, e.g., a person may smile without being happy.
In order to improve emotion recognition performance, \textit{multimodal learning} has been introduced. Most of the current research on multimodal emotion recognition focuses on the combination of video, audio, and language cues~\cite{delbrouck2020transformerbased,han2021bibimodal}. 
Such cues represent voluntary expressions of emotions, i.e., the \textit{behavioral} component. 
On the other hand, the \textit{physiological} component is largely involuntary and is directly activated when a person is emotionally triggered~\cite{shu2018review}. When an emotional trigger appears, a change in the physiological pattern is inevitable and detectable~\cite{jerritta2011physiological}; e.g., the heart rate increases if a person sees a snake. Thanks to their involuntary nature and to the ease of recording, peripheral signals as the electrocardiography (ECG) or photoplethysmogram (PPG), for the heart activity, and the electrodermal activity (EDA), for to the skin conductivity, have been exploited for the purpose of recognizing human emotions~\cite{kim2018stress, sanchezreolid2020machine, topic2022emotion}.

While unimodal video and physiological signals have been used to recognize emotions, very few works combine the two kind of data~\cite{siddharth2019utilizing, moin2023emotion, chaptoukaev2023stressid}. 
A recent work from 2023, proposes to fuse traditional features, classifying them by SVM and k-NN~\cite{moin2023emotion}. Similarly, in a recent dataset for stress identification~\cite{chaptoukaev2023stressid}, the baselines for both unimodal and multimodal experiments are conducted based on SVM/MLP using hand-crafted features for both video and physiological signals. 
The state-of-the-art approaches utilize old-fashioned machine learning pipelines. This is probably due to the small number and small size (max 60 subjects) datasets available.
We propose to fill this gap, designing a model that exploits the benefits of the transformer architecture \textit{attention}~\cite{vaswani2017attention}. The idea is to enable the use of long input sequences (1-2 minutes), in the context of \textit{emotion recognition}, with video and physiological data,
in order to extract complementary features.

\begin{figure}[t]
\centering
\includegraphics[width=\textwidth]{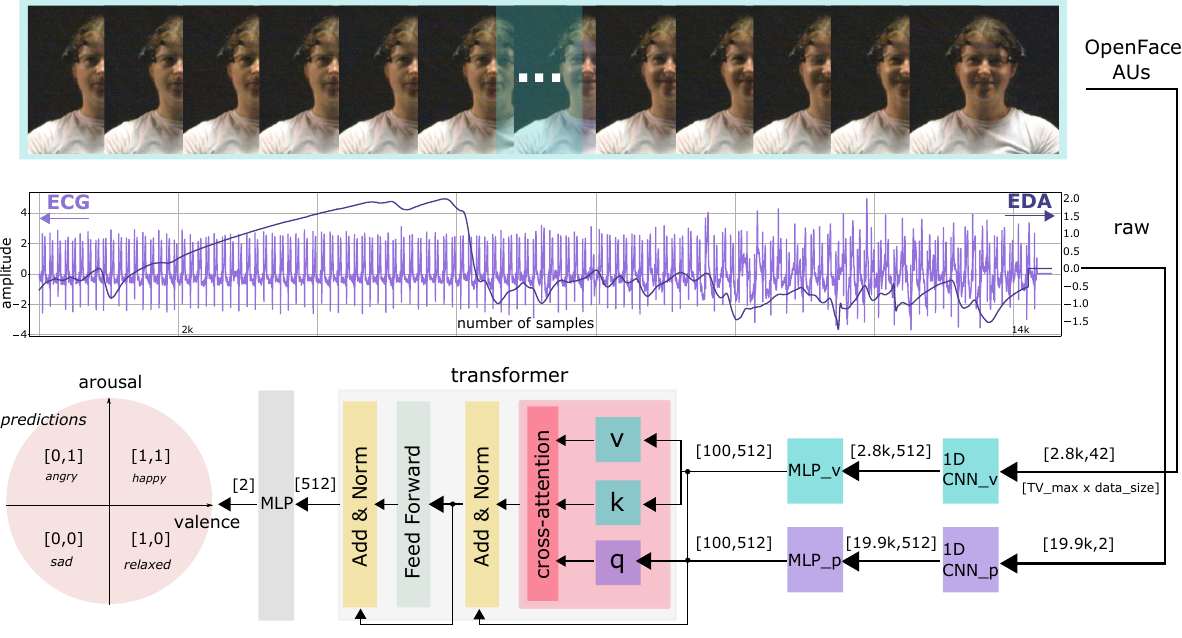}
\caption{Multimodal Emotion Recognition based on Video and
Physiological Signals (MVP) architecture. Video features and raw physiological signals are input as full long sequences into the model to predict binary valence and arousal. The cross-attention transformer is used to fuse multiple modalities.}
\label{fig:architecture}
\end{figure}

First, we study how to efficiently input the video data. We compare two approaches, the DL features extracted through VideoMAE~\cite{tong2022videomae} and the action units (AUs), extracted through OpenFace~\cite{baltrusaitis2018openface}.
VideoMAE and its modification~\cite{cai2023marlin} have just recently showed to achieve state-of-the-art results on multiple tasks, comprehending facial expression recognition. The main limitation of the available approaches is related to the input length. Typically short videos of few seconds are used. An open question is if and how these models can scale on longer videos. Here we adapt the Hugging Face implementation of VideoMAE to long (1-2 minutes) videos, and we compare these results with the traditional approach of AUs extraction. 
AUs are fine-grained facial muscle movements~\cite{ekman1978facial}, typically extracted through neural networks pre-trained on large sets of facial data.
Each AU is associated with a specific set of facial muscles.
As this is the first study on scaling VideoMAE for long emotion recognition videos, we assessed multiple configurations, on two datasets (AMIGOS and DEAP). 

Second, we study how to efficiently input physiological data. 
The state-of-the-art is based on 1D-CNN and unimodal transformer for the recognition of emotions from raw physiological data~\cite{vazquezrodriguez2022transformerbased, vazquez2022emotion}. 
One main limitation of these methods is related to the input signals, which are cut into 10-seconds segments. Here we improve the physiological backbone permitting the input of the full sequences, thus enabling attention to find correlation in both short and long range dependencies. 
Finally, we propose a multimodal transformer-based architecture, the Multimodal for Video and Physio (MVP), to efficiently couple the behavioral and physiological components for the emotion recognition task (Figure~\ref{fig:architecture}). The model combines facial video features with ECG/PPG and EDA. 
Deviating from other multimodal transformer-based architectures for video and language data~\cite{delbrouck2020transformerbased, han2021bibimodal, zhang2022transformerbased}, which use as input short subsequences (4-10 seconds), the MVP architecture is developed to handle long input sequences (1-2 minutes) of complementary behavioral and physiological data. Summarized contributions:
\begin{enumerate}
\item We comprehensively study the video and physiological backbones for inputting long sequences.
\item We propose MVP, a new affective computing architecture, streamlined to fuse video and physiological signals.
\item Our experimental results suggest that MVP outperforms former methods for emotion recognition based on facial videos, EDA, and ECG/PPG.
\end{enumerate}





\section{Related Work}

\begin{table}[ht]
\caption{State-of-the-art multimodal emotion recognition methods. V means visual, A means audio, L means language, P means physiological. 
We describe visual and physiological features in detail, but do not describe audio and linguistic features since our pipeline is based on video and physiological modalities. M-SENA is a platform with the state-of-the-art ER methods, from which we reported here the main characteristics of top three methods.}
\label{table:sota_mm}
\begin{center}
\begin{tabular}{c|c|c|c}
\toprule
Year & Framework & Modalities & Visual (V) and Physio (P) Features \\
\midrule
2019 & UDLBV~\cite{siddharth2019utilizing} & V, P & VGG-16 (V) \\
\hline
2020 & TBJE~\cite{zhang2022transformerbased} & V, A, L & R(2+1)D-152 outputs (V) \\
\hline
2021 & BBMF~\cite{han2021bibimodal} & V, A, L & Action Units (V) \\
\hline
& & & Action Units (V) \\
2022 & M-SENA~\cite{mao2022msena} & V, A, L & Facial Landmarks (V) \\
& & & Eye Gaze (V) \\
\hline
2022 & MultiTransformer~\cite{vazquez2022emotion} & P (ECG, EEG) & 1D-CNN + Transformer (P) \\
\hline
2023 & EHCI~\cite{moin2023emotion} & V, P & Hand-Crafted (V, P) \\
\hline
2023 & StressID~\cite{chaptoukaev2023stressid} & V, P (ECG, EDA) & Hand-Crafted (V, P) \\
\bottomrule
\end{tabular}
\end{center}
\end{table}

\subsection{Multimodal Emotion Recognition}
Table~\ref{table:sota_mm} shows the state-of-the-art emotion recognition methods. 
In the case of video, audio and language, transformer-based architectures have been proposed. The data are typically input into transformer as features~\cite{vazquezrodriguez2022transformerbased, zhang2022multitask, zhang2022transformerbased}: Action Units for the video (e.g., from OpenFace), audio features (e.g., Mel spectrogram from Librosa or Covarep), and language features (e.g., from GloVe or BERT). Each modality is used as input into self-attention layers, which allow the model to attend to different parts of the input sequence and capture long-range dependencies. Then the multimodal part takes place, performing cross-attention across modalities. The cross-attention can be done in multiple ways. The cross-modal fusion has been developed on pairwise representations~\cite{han2021bibimodal}. Here two couples of the modalities (i.e., language + audio and language + video, with language as the primary modality) enter a complementary module where bimodal fusion is done at the Multi-Head Attention level. Here, each modality has its own representation to preserve modality-specific features. 
A joint-encoding has been also used at the level of the Multi-Head Attention, to encode information from all the modalities, producing a unified representation~\cite{delbrouck2020transformerbased}. 
The best result is achieved with a bimodal model using the linguistic + acoustic cues. 
Notably, in both cases~\cite{delbrouck2020transformerbased,han2021bibimodal}, the language cue is the most representative. This can be due to the fact that annotators mainly worked on the transcript data. 
One of the major limitation of these works is related to the unique use of the behavioral component, i.e., video, audio and language data. 

Few attempts have been made to build a multimodal framework that can exploit both the behavioral and physiological components. These works rely on late fusion of classical features. In the UDLBV~\cite{siddharth2019utilizing} a CNN architecture (VGG-16 pretrained on ImageNet) and hand-crafted (HC) features are concatenated in a first step, after which PCA is applied, for then passing the features to an LSTM (only for the DEAP dataset) and finally classify with extreme learning machines (ELM). In EHCI~\cite{moin2023emotion} traditional features as spectral, HC, and local binary patterns for facial images are late fused and classified through SVM and k-NN. 

Our aim is to propose a transformer-based architecture that fuses video and physiological data through cross-attention. The idea is to improve the emotions recognition
by combining these two complementary signals, which represent the behavioral and the physiological components of emotions.

\subsubsection{Video cues.} State-of-the-art video-based emotion recognition methods use two main types of facial video features: (i) classical features like action units (AUs) and facial landmarks~\cite{han2021bibimodal,mao2022msena} and (ii) general features extracted through DL techniques~\cite{strizhkova2024maura, zhang2022multitask,zhang2022transformerbased}. We compare these two approaches: (i) the use of AUs, extracted through OpenFace~\cite{baltrusaitis2018openface}, and (ii) general features extracted through VideoMAE~\cite{tong2022videomae}. 
AUs are fine-grained facial muscle movements~\cite{ekman1978facial}, typically extracted through neural networks pre-trained on large sets of facial data. These methods have been largely used showing stable and reliable results. VideoMAE and its modifications~\cite{cai2023marlin, wang2023videomaev2} are new pre-training methods, providing the best results on many tasks, including video-based facial expression recognition. Masked auto-encoders (MAE) are self-supervised learners (SSL) trained to reconstruct the missing masked random patches of the input video. The use of VideoMAE in the emotion recognition field is very recent, therefore we report our adaptation to the challenging set-up of emotion recognition with long videos. 


\subsubsection{Physiological cues.} Peripheral physiological data are largely involuntary~\cite{shu2018review}, mediated by the autonomous nervous system, which is directly activated when a person is emotionally triggered. This means that when an emotional activation is present, a change in physiological patterns is inevitable and detectable~\cite{jerritta2011physiological}. 
Peripheral physiological data, as ECG/PPG and EDA, have been exploited to derive reliable information 
for emotion recognition~\cite{egger2019emotion}.
The use of machine learning on physiological data is quite recent, up to 5-6 years ago computational models were the most used. In the last years, classical machine learning pipelines, with models as Multilayer Perceptron (MLP), Support Vector Machine (SVM) and random forest trained on HC features~\cite{ross2021unsupervised} became more popular. 
The first approach exploiting deep learning architectures has seen the translation of physiological time-series data into 2D images, representing spectrogram or power spectral density, which are then input into CNN architectures (as ResNet or VGG) pretrained on different datasets (as ImageNet)~\cite{bagherzadeh2022recognition,elalamy2021multimodal,kahou2015recurrent}. GNNs have also been used~\cite{jia2021hetemotionnet} and two stacked convolutional autoencoders has been proposed to combine ECG and EDA~\cite{ross2021unsupervised}. Now, the state-of-the-art method in the field is based on 1D-CNN and self-attention transformers for the recognition of emotions from raw physiological data~\cite{vazquezrodriguez2022transformerbased, vazquez2022emotion}. The results have been proved only for the AMIGOS dataset. 
In ERMS~\cite{vazquez2022emotion} the two signals are used as time-series and the prediction is done after the concatenation of the last hidden layers of unimodal transformers. 
One main limitation of all these methods is related to the input signals, which are cut segments of 10-seconds. Then, or random 10s are taken for each sample, or each sample is used and the output of the features extractor is averaged. The advantage of using 10-seconds subsequences is that the network can work with a small and fixed input, resulting in a higher efficiency. 
This approach does not consider two factors. The ECG and EDA vary on different time scales, from few ms to tens of seconds/minute. The annotations are weak; a subject may express the labeled emotion for some seconds, few times in the whole recording, and being neutral the rest of the time. 
We advance the state-of-the-art permitting the input of the full sequences, thus enabling attention to find correlation in both short and long range dependencies. 

We compare our approach with with state-of-the-art methods for the input of physiological data~\cite{vazquezrodriguez2022transformerbased} and for the combination of multimodal signals~\cite{vazquez2022emotion}.

\section{Method}

\subsection{VideoMAE for Long Input Videos}

\begin{figure}[ht]
\centering
\includegraphics[width=.6\columnwidth]{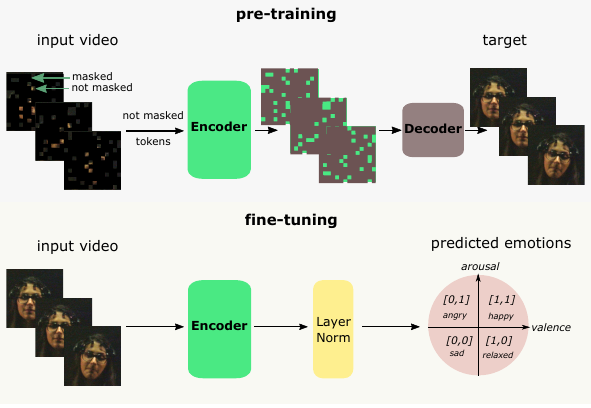}
\caption{Pre-training and fine-tuning steps of VideoMAE in emotion recognition. In the pre-training step the autoencoder reconstructs the masked input video. During the fine-tuning, the pre-trained encoder is fine-tuned to predict binarized valence and arousal from the original not masked videos.}
\label{fig:scaling}
\end{figure}

We use the Hugging Face implementation of VideoMAE, with the ViT-B encoder of 12 layers, hidden size of 768, and 87M parameters. During pre-training, the input video is masked with a ratio of 90\% and fed into the encoder, which outputs the latent representations. The shallow decoder then takes the latent representations from the encoder and reconstructs the input videos. During fine-tuning, the pre-trained encoder is learned to predict binarized valence and arousal from original (i.e. not masked) videos.
We perform experiments on two datasets, AMIGOS~\cite{mirandacorrea2021amigos} and DEAP~\cite{koelstra2011deap}.
We implemented and tested multiple configurations of pre-training strategy. (i) Self-supervised pre-training on Kinetics-400 and fine-tuning on the target dataset (i.e. AMIGOS or DEAP). (ii) Self-supervised pre-training on Kinetics-400, plus AMIGOS and DEAP, and fine-tuning on the target dataset (see Figure~\ref{fig:scaling}). (iii) Self-supervised pre-training on Kinetics-400, plus an intermediate step of supervised pre-training on Kinetics-400 and one of the two target datasets and fine-tuning on the other target dataset.
Table ~\ref{table:video_datasets} shows input video datasets used.
Figure~\ref{fig:scaling} represents the pipeline of VideoMAE pre-training, with masked input video and reconstruction task, and the fine-tuning step to predict valence and arousal.

\begin{table}[t]
\caption{Video datasets used in this study. KAD stays for Kinetics-400 \& AMIGOS \& DEAP.}
\label{table:video_datasets}
\centering
\begin{tabular}{cccc}
\toprule
Dataset & \# Video clips & Average clip duration & Source \\
\midrule
Kinetics-400 & 306245 & 10s & YouTube \\
AMIGOS & 640 & 80s & Lab \\
DEAP & 880 & 70s & Lab \\
\midrule
KAD & 307765 & - & Multi-source \\
\bottomrule
\end{tabular}
\end{table}

\subsection{Selected Backbones for Features Extraction}

\subsubsection{Vision backbone.}
Our video backbone is composed of $\text{1D-CNN}_v$ followed by $\text{MLP}_v$. It takes as input the AUs matrix of the full input video and outputs embeddings that are used as tokens in transformer.

\subsubsection{Physiological backbone.}
Our physiological backbone extracts temporal features from the entire raw physiological data using $\text{1D-CNN}_p$ and $\text{MLP}_p$ layer. 
We expanded the use of the sole $\text{1D-CNN}_p$~\cite{vazquezrodriguez2022transformerbased}, adding the $\text{MLP}_p$, to enable the handling of long sequences. 
The extracted features are used as tokens in transformer.

\subsection{MVP Architecture}
Figure~\ref{fig:architecture} shows the MVP architecture. 
Here, one input trial is exemplified, consisting of a video sequence $SV_{ij}$ and a physiological sequence $SP_{ij}$, of shapes $TV_{max} \times 42$ and $TP_{max} \times 2$, respectively.
In our experiments, we use a batch size of 8, so at each training iteration we randomly choose 8 trials and perform all the calculations for this batch.
The input data enters the video and physiological backbones.
For the video input, the zero-padded matrix with AUs enters $\text{1D-CNN}_v$ followed by $\text{MLP}_v$. For the physiological data, the zero-padded matrix with concatenated ECG and EDA enters $\text{1D-CNN}_p$ followed by $\text{MLP}_p$. 
$\text{1D-CNN}_v$ and $\text{1D-CNN}_p$ extract features from input data without changing the time dimension.
$\text{MLP}_v$ and $\text{MLP}_p$ reduce the time dimension without changing the feature dimension.
$\text{1D-CNN}_p$ takes as input a raw physiological sequence of dimension [19.9k, 2] and outputs a feature sequence of dimension [19.9k, 512].
Then $\text{MLP}_p$ takes as input the output from $\text{1D-CNN}_p$ and outputs a compact representation [100, 512].

After that, the outputs from 1D-CNNs are used as input into transformer with cross-attention for fusing video with physiological signals. The MVP method uses mid-fusion fusion, where video modality is input as keys (k) and values (v) and physiological one as queries (q) (Equation~\ref{eqn:crossatt}).
\begin{equation}
\label{eqn:crossatt}
\begin{aligned}
\text{Attention}(Q_p,K_v,V_v) = \text{Softmax}\left(\frac{Q_pK_v^T}{\sqrt{d_k}}\right) \cdot V_v
\end{aligned}
\end{equation}
where $Q_p$ is the query from the physiological data and $K_v$ and $V_v$ are the keys and the values from the video data.
transformer has 8 attention heads and 8 attention layers and it predicts two outputs: valence and arousal, binarized, as reported in Figure~\ref{fig:architecture}.
We use binary cross entropy (BCE) to calculate the loss.
During training the loss is backpropagated through the neural network layers to calculate the gradient for each weight. 

\subsection{Data Handling}

The entire sequence of video and physiological data is used, with a length that can vary between 60s and 155s.
As the adopted datasets contain multiple trials from different subjects, the input data for a trail $i$ of a subject $j$ 
is composed of two matrices: $SV_{ij}$ for video and $SP_{ij}$ for physiological signals, with dimensions $[TV_{ij}, 42]$ and $[TP_{ij}, 2]$, respectively, where $TV_{ij}$ and $TP_{ij}$ correspond to lengths of the sequences and $42$ and $2$ are input data dimensions of each modality.
Video and physiological signals are sampled at different rates, 
therefore $TV_{ij}$ and $TP_{ij}$ are different (e.g., for the AMIGOS dataset $TV_{max}$ is 2.8k and $TP_{max}$ is 19.9k, corresponding to~155s sampled at 18fps and 128Hz, for the video and the physiological data, respectively).
We extract 42 visual features from each video frame and use the sequence of AUs and eye gaze as video representation.
AUs are fine-grained facial muscle movements~\cite{ekman1978facial}, each of which relates to a subset of extracted facial landmarks~\cite{perveen2020configural}. From each frame we extract the following AUs: $1, 2, 4, 5, 6, 7, 9, 10, 12, 14, 15, 17, 20, 23, 25, 26, 28$, and $45$, by means of the OpenFace library~\cite{baltrusaitis2018openface}.
Each AU is described in two ways: presence, if AU is visible in the face, and intensity, how intense is the AU on a 5-point scale.
The eye gaze corresponds to two gaze direction vectors of each eye.
The physiological data has a dimension of $2$ at each timestamp, corresponding to the $2D$ time series, made of ECG and EDA raw signals. 

Before using the data as input to our model, we perform the following pre-processing.
We find the longest sequence and use its $TV_{max}$ and $TP_{max}$ lengths to construct input matrices for all the trials in the dataset. 
Each pair of input sequences $SV_{ij}$, $SP_{ij}$ is zero-padded so that all input matrices with AUs and physiological signals have sizes $TV_{max} \times 42$ and $TP_{max} \times 2$, respectively. 

\begin{figure}[t]
\centering
\includegraphics[width=.8\columnwidth]{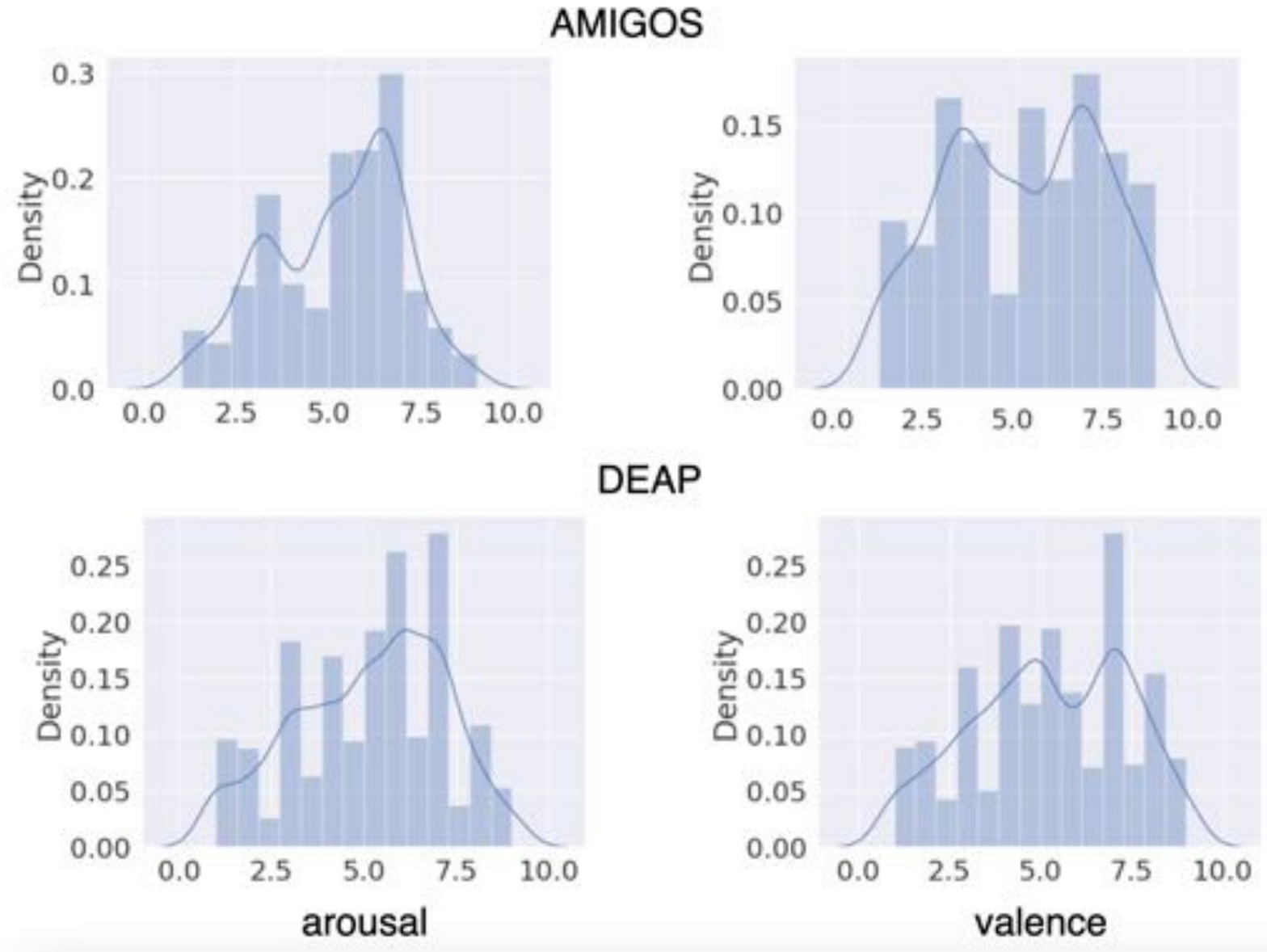}
\caption{Label distribution of AMIGOS and DEAP datasets.}
\label{fig:label}
\end{figure}

\section{Experiments}

\subsection{Emotion Recognition Labels}
Emotion classification is typically approached by measuring arousal and valence based on the emotion circumplex model~\cite{russell1980circumplex}. Valence refers to the degree of positive or negative affect associated with an emotion. Arousal refers to the degree of activation, engagement associated with an emotion. The valence and arousal dimensions have been used to categorize emotions into quadrants (see Figure~\ref{fig:architecture}): high valence/high arousal (e.g., excitement, happiness), high valence/low arousal (e.g., relaxation), low valence/high arousal (e.g., anger), and low valence/low arousal (e.g., boredom, tiredness). In the classification task these labels are typically extracted by the Self-Assessment Manikin (SAM) scale reported by each subject. SAM is a pictorial tool for rating valence and arousal on a scale of 1 to 9. A subject selects the pictogram that best corresponds to its emotional state at the end of each elicitation video. Therefore, such labels are more subjective and referred to the whole session, that typically last between 1 and 2 minutes.
In Figure~\ref{fig:label}, the label distribution of the used datasets is reported.

\subsection{Datasets and Pre-processing}
The AMIGOS~\cite{mirandacorrea2021amigos} and the DEAP~\cite{koelstra2011deap} datasets are available online and comprise both video and physiological data. The recordings are performed in a laboratory setting and the participants are instructed to watch emotional videos and then respond to the Self-Assessment Manikin scale. 
The AMIGOS dataset includes the recording of 40 subjects, who rated 16 movie video clips for valence, arousal, and other measures (i.e., dominance, liking, familiarity and basic emotions).
The DEAP dataset includes the recording of 32 subjects, who rated 40 music video clips for valence, arousal and other measures.
We use the facial videos and EDA + ECG for AMIGOS, and the facial videos and EDA + PPG for DEAP, preprocessed as follow. 

\begin{figure}[t]
\centering
\includegraphics[width=1\columnwidth]{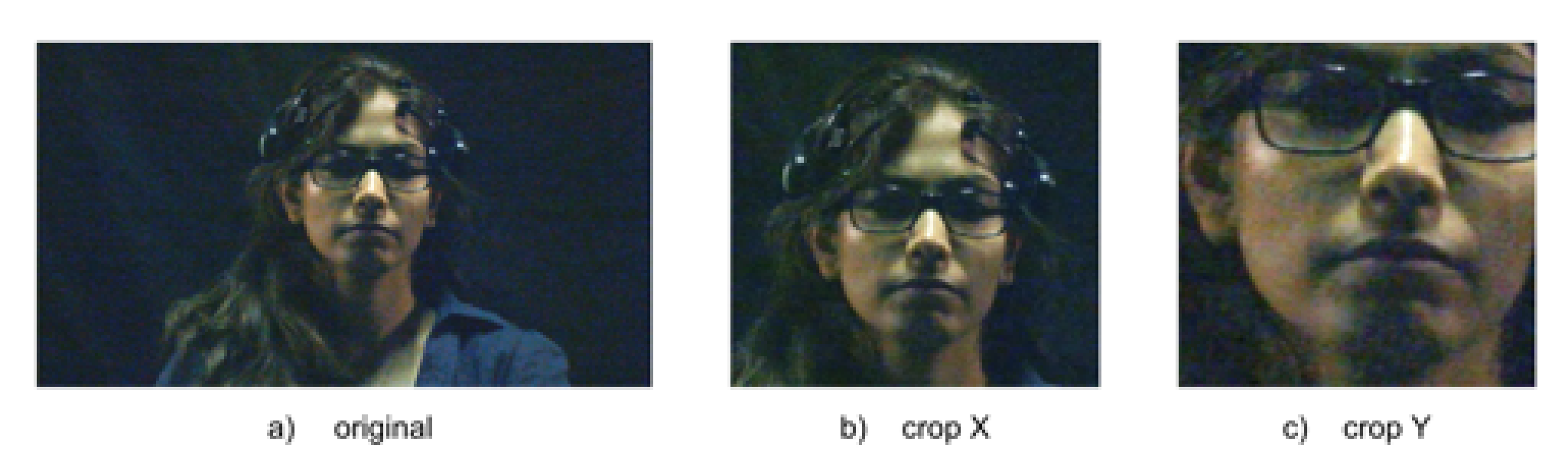}
\caption{Crop comparison on the Amigos dataset. The larger crop X captures the entire face and a bit of background, the smaller crop Y captures exclusively the face. Larger crops give higher F1-score.}
\label{fig:crop}
\end{figure}

For VideoMAE, the input videos are cut into smaller clips of 4 seconds, from which 16 frames are taken and cropped using facial landmarks generated by OpenFace~\cite{baltruvsaitis2016openface}. 
Two different cropping strategies were evaluated (Figure~\ref{fig:crop}). The larger one captures the entire face and a bit of background, ensuring the face remains visible even when the subject head is in motion, the smaller one captures exclusively the face, causing occasionally and partially cropping out. 
We experimented the two cropping strategies, achieving higher performances (7-9\%) with the larger crops; thereafter this is the set-up adopted in this work. 

For the AUs, we use OpenFace~\cite{amos2016openface} to extract AUs, as it is a stable and largely adopted library in the community. 42 AUs and eye gaze direction features are extracted for each frame.

For the physiological data, we downsampled to 128Hz, then we filtered the signals with Butterworth filter 
plus powerline filter (to remove the 50Hz). The Butterworth filters used are: 3th order band-pass with [0.5,8]Hz band for PPG, 5th order high-pass with a 0.5Hz cutoff for ECG, 4th order low-pass with a 3Hz cutoff for EDA.
In the AMIGOS dataset, we cut the first 1s in all the physiological signals, as artefacts are present in most of the recordings (probably due to the device connection). 
Since the signal sequences have different lengths, we pad the shorter sequences with zeros, so that all input sequences have the same length. 
The raw data is normalized with zero-mean and unity-variance before entering the pipeline. 

\subsection{Experimental Set-up}

We use 5-fold cross validation to split data into train and test subject independently. All trials of one subject are either in the train or in the test split.
We use the weighted F1-scores as metric, as the datasets are imbalanced, reporting the mean and standard deviation. 

We split valence and arousal continuous labels using the thresholds of 4.5 and 5 for AMIGOS and DEAP, respectively. If the continuous label is less or equal than the threshold, the binarized label is 0 and 1 otherwise. 

\section{Results and Discussions} 

\subsection{Input Features}
Our model fuses video with physiological signals.
To extract features from each modality, we use the models that best fit the input data.
Here we explain our motivation in selecting feature extraction backbones for facial videos and physiological signals (ECG and EDA).

\subsubsection{VideoMAE for emotion recognition.}
The results in Tables~\ref{table:videomae-amigos} and~\ref{table:videomae-deap} are related to the valence prediction for the AMIGOS and DEAP datasets using the VideoMAE self-supervised pre-training. The results show that the most effective approach is the three steps strategy, where, after the SSL pre-training on Kinetics-400, a supervised pre-training step is added before fine-tuning on the target dataset. Kinetics-400 is biggest dataset typically used to pre-train VideoMAE. This is related to the need of having data with the same distribution in the pre-training step. 

While increasing the amount of data in the self-supervised pre-training phase does not achieve the highest weighted F1-score, it is still more effective than solely relying on Kinetics-400 for supervised pre-training or not using it at all. Additionally, when the supervised pre-training step is added, the model adapts faster. For instance, with DEAP, the model requires only 3 epochs to fine-tune, as opposed to the initial 10. This further supports the intuition that leveraging datasets from the same domain can significantly benefit the learning process.


\begin{table}[t]
\caption{Valence prediction from raw videos using self-supervised VideoMAE pre-training and supervised pre-training. KAD is Kinetics-400 \& AMIGOS \& DEAP.}
\begin{subtable}{1\textwidth}
\centering
\begin{tabular}{c|c|c}
\toprule
SSL pre-training data & Supervised pre-training data & F1-score weighted \\
\midrule
Kinetics-400 & AMIGOS & 49 $\pm$ 5 \\
KAD & AMIGOS & 50 $\pm$ 4\\
Kinetics-400 & Kinetics-400 \& AMIGOS & 51 $\pm$ 6 \\
Kinetics-400 & KAD & \textbf{59 $\pm$ 5} \\
\bottomrule
\end{tabular}
\caption{5-fold cross validated F1-score for the AMIGOS dataset.}
\label{table:videomae-amigos}
\end{subtable}
\bigskip
\begin{subtable}{1\textwidth}
\centering
\begin{tabular}{c|c|c}
\toprule
SSL pre-training data & Supervised pre-training data & F1-score weighted\\
\midrule
Kinetics-400 & DEAP & 52 $\pm$ 4 \\
KAD & DEAP & 55 $\pm$ 7\\
Kinetics-400 & Kinetics-400 \& DEAP& 53 $\pm$ 5 \\
Kinetics-400 & KAD & \textbf{60 $\pm$ 6} \\
\bottomrule
\end{tabular}
\caption{5-fold cross validated F1-score for the DEAP dataset.}
\label{table:videomae-deap}
\end{subtable}
\end{table}

\subsubsection{Selection of the video backbone.}
We compare two approaches: (i) the use of AUs extracted through OpenFace, and (ii) DL features extracted through VideoMAE~\cite{tong2022videomae}.
We compare the results on the AMIGOS dataset in Table~\ref{table:amigos_main}. 
The two approaches give similar results, with AUs outperforming VideoMAE features.
We selected AUs features as they give higher F1-score, are more reliable, explainable and requires less memory for training. A possible explanation for their performance is that the model used to extract AUs is pre-trained on a large set of face data, so the predictions incorporate extra knowledge, providing stable results. In this context, stable means that the extracted features are meaningful under different lighting conditions, resolutions, and head angles. Indeed, on the AMIGOS dataset, where the videos are quite dark and low resolution, AUs improve the results even more.

\begin{table}[t]
\caption{Comparison of state-of-the-art emotion recognition methods with the proposed MVP for unimodal and multimodal emotion recognition. ECG+EDA means concatenated raw ECG and EDA used as one modality. Video input is raw videos for VideoMAE and AUs for other methods. HC means hand-crafted.}
\begin{subtable}{1\textwidth}
\centering
\begin{tabular}{c|cc|cc|cc}
\toprule
\multirow{3}{*}{Method} & \multicolumn{6}{c}{F1-score weighted} \\ 
\cline{2-7} & \multicolumn{2}{c|}{ECG+EDA} & \multicolumn{2}{c|}{video} & \multicolumn{2}{c}{ECG+EDA+video} \\
\cline{2-7} & arousal & valence & arousal & valence & arousal & valence \\
\midrule
SVM (HC feat) \cite{chaptoukaev2023stressid} & 51 $\pm$ 5 & 60 $\pm$ 6 & 44 $\pm$ 8 & 59 $\pm$ 3 & 49 $\pm$ 8 & 64 $\pm$ 3 \\
MLP (HC feat) \cite{chaptoukaev2023stressid} & 57 $\pm$ 4 & 63 $\pm$ 1 & 53 $\pm$ 5 & 64 $\pm$ 5 & 55 $\pm$ 5 & 63 $\pm$ 3 \\
UniTransformer (adapted) \cite{vazquezrodriguez2022transformerbased} & 54 $\pm$ 4 & 59 $\pm$ 4 & --- & --- & --- & --- \\
MultiTransformer (adapted) \cite{vazquez2022emotion} & --- & --- & --- & --- & 56 $\pm$ 6 & 61 $\pm$ 3 \\
VideoMAE \cite{tong2022videomae} & --- & --- & 53 $\pm$ 6 & 59 $\pm$ 5 & --- & --- \\
\midrule
UniAUTransformer (ours) & --- & --- & 56 $\pm$ 3 & 63 $\pm$ 4 & --- & --- \\
UniPhysioTransformer (ours) & 56 $\pm$ 3 & 62 $\pm$ 4 & --- & --- & --- & --- \\
MVP (10s cut input, ours) & --- & --- & --- & --- & 55 $\pm$ 5 & 60 $\pm$ 6 \\
MVP (ours) & --- & --- & --- & --- & \textbf{58 $\pm$ 6} & \textbf{66 $\pm$ 4} \\
\bottomrule
\end{tabular}
\caption{5-fold cross validated F1-score for the AMIGOS dataset.}
\label{table:amigos_main}
\end{subtable}
\bigskip
\begin{subtable}{1\textwidth}
\centering
\begin{tabular}{c|cc|cc|cc}
\toprule
\multirow{3}{*}{Method} & \multicolumn{6}{c}{F1-score weighted} \\ 
\cline{2-7} & \multicolumn{2}{c|}{ECG+EDA} & \multicolumn{2}{c|}{video} & \multicolumn{2}{c}{ECG+EDA+video} \\
\cline{2-7} & arousal & valence & arousal & valence & arousal & valence \\
\midrule
MLP (HC feat) \cite{chaptoukaev2023stressid} & 52 $\pm$ 3 & 53 $\pm$ 2 & 51 $\pm$ 2 & 54 $\pm$ 3 & 52 $\pm$ 2 & 55 $\pm$ 6 \\
VideoMAE \cite{tong2022videomae} & --- & --- & 52 $\pm$ 3 & 60 $\pm$ 6 & --- & --- \\
\hline
MVP (ours) & 53 $\pm$ 6 & 54 $\pm$ 4 & 54 $\pm$ 4 & 57 $\pm$ 7 & \textbf{ 55 $\pm$ 5 } & \textbf{61 $\pm$ 4} \\
\bottomrule
\end{tabular}
\caption{5-fold cross validated F1-score for the DEAP dataset.}
\label{table:deap_main}
\end{subtable}
\end{table}

\subsubsection{Selection of the physiological backbone.}
We compare the state-of-the-art unimodal transformer for physiological data~\cite{vazquezrodriguez2022transformerbased} and our physiological backbone. The unimodal transformer backbone uses 10s subsequences and 1D-CNN while our backbone is made of 1D-CNN + MLP and data is input as whole sequences. The experiments are run on the ECG signal of the AMIGOS dataset. 
Table \ref{table:amigos_main} shows that our proposed approach outperforms the state-of-the-art by around 3 percentage points in predicting valence. This means that a deeper temporal understanding is beneficial for the valence.
\subsection{Multimodal Emotion Recognition}

The need of combining video and physiological data through a novel DL architecture is investigated. We compare the MVP method with classical machine learning approaches and more recent transformer-based architectures proposed for both unimodal and multimodal data. 
The classical machine learning pipeline used here exploits hand-crafted (HC) features as input to Multi-Layer Perceptron
(MLP) and Support Vector Machine (SVM) to classify valence and arousal \cite{chaptoukaev2023stressid}. All transformer models are used with adaptations to ensure fair comparison. In UniTransformer we combine ECG and EDA, whereas the original UniTransformer uses only ECG as input. In MultiTransformer, we add AUs as video component, using ECG+EDA as one input and AUs as the second one. The original MultiTransformer uses ECG and EEG. The main difference between MultiTransformer and MVP is the fusion. In MultiTransformer late fusion is used, while MVP uses mid-fusion. For VideoMAE we have studied and adapted the method to the case of long facial videos. Tables~\ref{table:amigos_main} and~\ref{table:amigos_main} show that transformers outperform classical machine learning. And the use of mid-fusion and a temporal model able to input the full sequence, as is done in MVP, gives the best results. This is due to the capability of cross-attention to find long-range dependencies between different kinds of input data.

\subsection{Success Case Analysis}
\begin{figure}[t]
\centering
\begin{subfigure}{.5\textwidth}
\centering
\includegraphics[width=.95\linewidth]{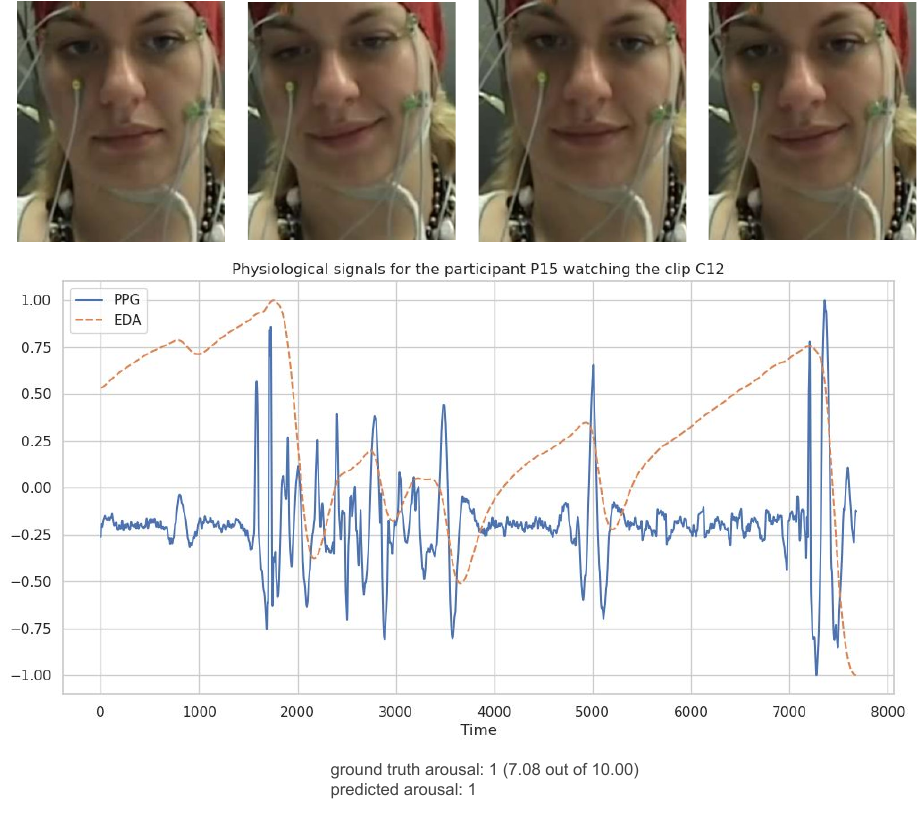}
\caption{The success case for the arousal prediction.}
\label{fig:success_arousal}
\end{subfigure}%
\begin{subfigure}{.5\textwidth}
\centering
\includegraphics[width=.95\linewidth]{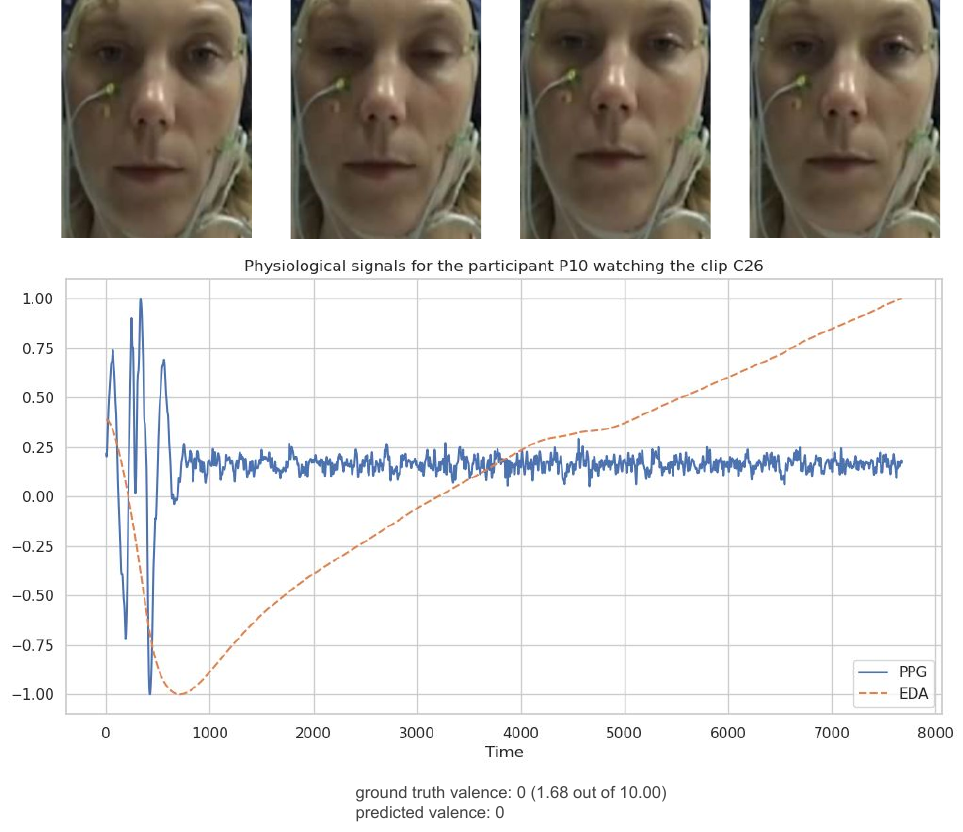}
\caption{The success case for the valence prediction.}
\label{fig:success_valence}
\end{subfigure}
\caption{The success cases for the arousal and valence prediction on the DEAP dataset.}
\label{fig:success_deap}
\end{figure}

We conduct case studies to show videos and physiological sequences in which the proposed MVP method outperforms state-of-the-art competitors. We compare our MVP with MLP proposed by StressID for multimodal fusion~\cite{chaptoukaev2023stressid}.
Table~\ref{table:deap_main} shows that MVP outperforms MLP on the DEAP dataset. Figures~\ref{fig:success_arousal} and~\ref{fig:success_valence} demonstrate the sequences where MVP gives better results than the competitor. We show both physiological signals and video frames of our success cases. 
Figure~\ref{fig:success_arousal} shows the case where the true arousal is 1, our MVP method predicts 1, and the competitor predicts 0. The video frames above show that the participant is initially neutral, and after a few seconds they smile, while at the same time the physiological signals change. The result demonstrates that MVP better captures the dynamics of the sequence.
Figure~\ref{fig:success_valence} shows the case where the true valence is 0, our MVP method predicts 0, and the competitor predicts 1. The video frames above show that the participant remains neutral throughout the experiment, without any changes except blinking while the physiological signal changes. This success case shows that MVP better fuses multiple modalities than MLP.

\section{Conclusions and Future Work}
We propose a new method for combining complementary data, the behavioral and physiological component of emotions. The method relies on the relevance of inputting long full sequences, exploiting attention. Regarding the video input, AUs are performing better than deep learning features in extracting facial representations for emotion recognition. Nevertheless, a strategy that includes similar data distribution in the pre-training step shows improved results for deep learning features.
Our experimental results showcase that fusing video and physiological signals outperforms each modality individually. 
MVP achieves state-of-the-art results in the challenging field of multimodal emotion recognition with video and physiological data, where small datasets are available. 
With this work we aim to foster research in the field, to improve the understanding of human emotions expressed by behavioral and physiological signals. 

Future works involve exploiting ECG and EDA data as two separated modalities, fused by a transformer model. We will develop a dedicated pre-training technique, in order to increase recognition capabilities of the model and to bring insight into physiological and behavioral modalities.

\section*{Acknowledgements}
This work was supported by European Union under Horizon Europe project GAIN (GA \#101078950), by French National Research Agency under UCA\textsuperscript{JEDI} Investments into the Future (ANR-15-IDEX-01), and by 3IA Côte d'Azur.



%
%
\bibliographystyle{splncs04}
\bibliography{main}
\end{document}